\pdfoutput=1

\documentclass[11pt]{article}

\usepackage[final]{acl}

\usepackage{times}
\usepackage{latexsym}

\usepackage{amsmath}

\usepackage[skins]{tcolorbox}

\usepackage[T1]{fontenc}

\usepackage[utf8]{inputenc}

\usepackage{microtype}

\usepackage{inconsolata}

\usepackage{graphicx}

\usepackage{arydshln}
\usepackage{amssymb}
\usepackage{pifont}
\usepackage{xcolor}
\usepackage{color,soul}

\usepackage{subfig}

\definecolor{myblue}{RGB}{65, 100, 166}
\definecolor{mycyan}{RGB}{75, 186, 186}

\newtcolorbox{instructionframe}[2][]{%
  enhanced,colback=white,colframe=myblue,coltitle=white,boxrule=1.0pt,
  fonttitle=\mdseries,
  attach boxed title to top left={yshift=-0.5\baselineskip-0.4pt,xshift=2mm},
  boxed title style={tile,size=minimal,left=1.5mm,right=1.5mm,
    colback=myblue,before upper=\strut},
  title=#2,#1
}

\title{MediQAl: A French Medical Question Answering Dataset for Knowledge and Reasoning Evaluation}

\author{Adrien Bazoge$^{1,2}$ \\
  $^1$Data Clinic, University Hospital of Nantes, France\\
  $^2$Nantes Université, École Centrale Nantes, CNRS, LS2N, France\\
  \texttt{adrien.bazoge@univ-nantes.fr}}

\begin{document}
\maketitle
\begin{abstract}

This work introduces MediQAl, a French medical question answering dataset designed to evaluate the capabilities of language models in factual medical recall and reasoning over real-world clinical scenarios. MediQAl contains 32,603 questions sourced from French medical examinations across 41 medical subjects. The dataset includes three tasks: (i) Multiple-Choice Question with Unique answer, (ii) Multiple-Choice Question with Multiple answer, and (iii) Open-Ended Question with Short-Answer. Each question is labeled as \textit{Understanding} or \textit{Reasoning}, enabling a detailed analysis of models' cognitive capabilities. We validate the MediQAl dataset through extensive evaluation with 14 large language models, including recent reasoning-augmented models, and observe a significant performance gap between factual recall and reasoning tasks. Our evaluation provides a comprehensive benchmark for assessing language models' performance on French medical question answering, addressing a crucial gap in multilingual resources for the medical domain.
\end{abstract}

\section{Introduction}

Medical licensing examinations, originally designed to assess students' knowledge and reasoning, are increasingly repurposed as benchmarks for evaluating large language models (LLMs) medical capabilities~\cite{yan2024largelanguagemodelbenchmarks}. Benchmarks of question-answering tasks predominantly rely on multiple-choice questions (MCQs) with a single correct answer~\cite{hendrycks2021measuring, wang2024mmlupro}. This format is widely used due to the availability of automatic evaluation metrics that provide consistent and objective assessment of LLMs at scale.
While MCQ-based datasets and metrics provide a valuable initial insight into LLM performance, they are often limited in several key aspects: the number of examples, the diversity of medical subjects covered, their representation of real-world clinical scenarios~\cite{shi-etal-2024-medical}, and the range of languages represented. Indeed, most of existing benchmarks are heavily centered around English, which restricts their applicability to multilingual or non-English contexts~\cite{yan2024largelanguagemodelbenchmarks}. Furthermore, medical benchmarks inherently reflect cultural, educational, and regulatory contexts in which they are developed. The format of questions and answers mirrors how medicine is taught and assessed in their respective countries, which differs in structure, emphasis, and evaluative expectations across regions. In addition, treatment guidelines, clinical protocols, and legal standards are often country-specific, meaning that identical questions translated across languages can pose entirely different challenges.

Recent efforts have focused on increasing diversity in question difficulty and covering a wider variety of medical subjects~\cite{zuo2025medxpertqabenchmarkingexpertlevelmedical}, yet these benchmarks remain predominantly limited to English-language and rely on MCQs with a single correct answer. This limitation is particularly problematic, as several studies have demonstrated significant performance disparities between languages, with LLMs performing considerably better in English compared to less-resourced languages~\cite{10.1145/3589334.3645643, dey2024betteraskenglishevaluation, ALONSO2024102938}. Therefore, it is crucial to develop more inclusive benchmarks that cover a broader range of languages and are more reflective of real-world clinical scenarios, ensuring a fair and comprehensive evaluation of LLMs in the medical domain.

In this work, we present MediQAl, a medical question answering dataset for French. This dataset contains questions sourced from French medical licensing examinations. These are manually created by academic and hospital faculty members to reflect real-world clinical scenarios and cover a broad range of medical subjects.

\begin{table*}[!htb]
\setlength\tabcolsep{4pt}
\centering
\resizebox{2\columnwidth}{!}{%
\begin{tabular}{l|ccc|ccc|ccc}
\hline
                                            & \multicolumn{3}{c|}{\textbf{MCQU}} & \multicolumn{3}{c|}{\textbf{MCQM}} & \multicolumn{3}{c}{\textbf{OEQ}} \\ \hline
                                            & \textit{Understanding}      & \textit{Reasoning}       & \textit{Total}     & \textit{Understanding}      & \textit{Reasoning}      & \textit{Total}      & \textit{Understanding}      & \textit{Reasoning}      & \textit{Total}      \\ \hdashline
\textbf{Total Number of Questions}                          & 11,336  & 5,681    & 17,017 & 7,742   & 2,875   & 10,617  & 1,842   & 3,125   & 4,969   \\
\multicolumn{1}{r|}{\textbf{\# Isolated Questions}}   & 9,126   & 961     & 10,087 & 6,200   & 343    & 6,543   & 836    & 179    & 1,015   \\
\multicolumn{1}{r|}{\textbf{\# In-context Questions}} & 2,210   & 4,720    & 6,930  & 1,542   & 2,532   & 4,074   & 1,006   & 2,946   & 3,954   \\
\textbf{Avg Question Length}                         & 18.95  & 21.57   & 19.82 & 13.20  & 16.12  & 13.99  & 16.79  & 20.95  & 19.40  \\
\textbf{Avg Clinical Scenarios Length}               & 83.50  & 107.67  & 99.97 & 94.87  & 114.77 & 107.24 & 109.71 & 141.28 & 132.19 \\
\textbf{Avg Answer Length}                           & -      & -       & -     & -      & -      & -      & 25.26  & 40.24  & 34.68 \\ \hline
\end{tabular}%
}
\caption{Characteristics of the MediQAl dataset. \textit{In-context questions} refers to questions including a clinical scenario. Lengths are measured in words.}
\label{tab:MediQAl_carac}
\end{table*}

This paper makes the following contributions:
\begin{enumerate}
\item We introduce MediQAl, a French medical question answering (QA) dataset that includes three tasks : (i) Multiple-Choice Question with Unique Answer (MCQU), (ii) Multiple-Choice Question with Multiple Answers (MCQM) and (iii) Open-Ended Question with Short-Answer (OEQ). 

\item MediQAl covers a total of 41 medical subjects and each question is categorized as either \textit{Understanding} or \textit{Reasoning}, enabling detailed analysis of LLMs' capabilities across different cognitive tasks.

\item We present an extensive evaluation of 14 large language models (LLMs) on MediQAl, including latest reasoning-based models, providing a comprehensive benchmark for assessing their performance over real-world clinical scenarios.  We compare different groups of models, focusing on the performance gap between vanilla models (non-reasoning) and their reasoning-enhanced counterparts.

\end{enumerate}

The dataset is available on HuggingFace\footnote{\href{https://huggingface.co/datasets/ANR-MALADES/MediQAl}{https://huggingface.co/datasets/ANR-MALADES/MediQAl}} under CC-BY-4.0 license and all evaluation scripts are available on Github\footnote{\href{https://github.com/abazoge/MediQAl}{https://github.com/abazoge/MediQAl}}.

\section{Related Work}

Multiple-choice question answering with a unique correct answer (MCQU) is a well-established task, frequently used to benchmark language models. This task is particularly prominent in the medical domain, where several datasets have been developed in various languages.
In English, multiple high-quality datasets exist, including HEAD-QA~\cite{vilares-gomez-rodriguez-2019-head}, MedQA~\cite{app11146421}, MedMCQA~\cite{pmlr-v174-pal22a}, MMLU (Medical) ~\cite{hendrycks2021measuring, wang2024mmlupro}, and more recently, MedXpertQA~\cite{zuo2025medxpertqabenchmarkingexpertlevelmedical}, an expert-level benchmark for medical MCQU tasks.

In other languages, efforts have been made to extend the task to non-English settings. Notable examples include datasets for Chinese~\cite{li-etal-2021-mlec}, Polish~\cite{bean2024do} and Spanish~\cite{ALONSO2024102938}. However, for French, the resources remain scarce. FrenchMedMCQA~\cite{labrak-etal-2022-frenchmedmcqa} is a dataset containing 3,105 multiple-choice questions, with both unique and multiple answers, but is limited to pharmacy topics. Another MCQU dataset, MedExpQA~\cite{ALONSO2024102938}, includes a French subset that is translated from Spanish.

In addition to multiple-choice datasets, open-ended question answering in the medical domain is less common, as evaluating free-text responses is more challenging and often requires manual human validation. Some open-ended QA datasets are derived from existing multiple-choice corpora. For instance, MEDQA-OPEN~\cite{nachane2024shotchainofthoughtdrivenreasoning} reformulates MedQA questions into an open-ended format. For French, there is only a single small-scale open-ended QA dataset, MedFrenchmark~\cite{Quercia2024-qy}, containing only 114 examples.

\section{MediQAl}

We introduce MediQAl, a French medical dataset consisting of questions sourced from French medical examinations. MediQAl is designed to evaluate medical knowledge and reasoning on both isolated and in-context questions reflecting real-world clinical scenarios. The dataset includes three subsets, corresponding to distinct question answering tasks: (1) \textit{Multiple-Choice Question with unique answer} (MCQU), (2) \textit{Multiple-Choice Question with multiple answers} (MCQM) and (3) \textit{Open-Ended Question with a short answer} (OEQ). MediQAl contains a total of 32,603 questions, of which 17,017 are MCQU, 10,617 are MCQM and 4,969 are OEQ. These questions span 41 medical subjects and are categorized as \textit{Understanding} or \textit{Reasoning}, offering a diverse and reliable benchmark for medical question answering tasks in French. Table~\ref{tab:MediQAl_carac} summarizes the main characteristics of the dataset.

\begin{figure*}[!ht]
\centering
\includegraphics[width=\textwidth]{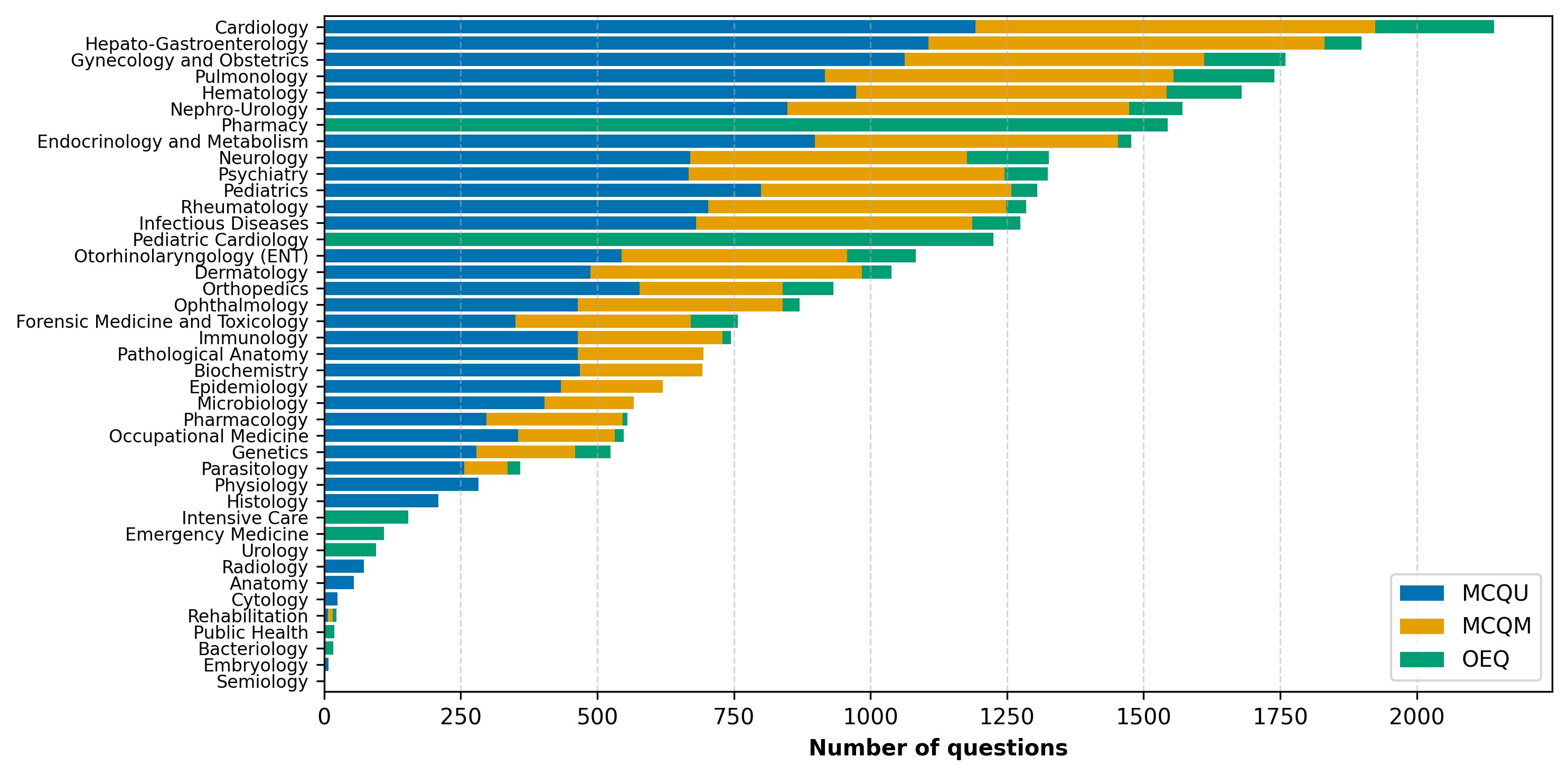}
\caption{Distribution of medical subjects across MediQAl dataset.}
\label{fig:figure_medical_subject}
\end{figure*}

\subsection{Tasks Definition} 

\paragraph{Multiple-Choice Question with Unique Answer (MCQU)} This task can be formulated as $X = \{C, Q, (O_1, ..., O_5), A\} $ where $C$ is an optional clinical scenario, $Q$ is the question, $(O_1, ..., O_5)$ are five candidate options and $A$ is the correct answer. For a given triplet $\{C, Q, (O_1, ..., O_5)\} $, the correct answer $A$ is a single option $O_i$ from $(O_1, ..., O_5)$. This task is similar to most existing MCQA datasets.

\paragraph{Multiple-Choice Question with Multiple Answers (MCQM)} This task follows a similar formulation as MCQU: $X = \{C, Q, (O_1, ..., O_5), A\} $. However, in MCQM, the correct answer $A$ is a subset of candidate options with $|\mathcal{A}| \geq 2$. The answer includes multiple correct options among $(O_1, ..., O_5)$.

\paragraph{Open-Ended Question with Short Answer (OEQ)} The OEQ task can be formulated as $X = \{C, Q, A\}$ where $C$ is an optional clinical scenario, $Q$ is the question and $A$ is a short, free-text answer. The answer length is lower than 200 tokens.

\subsection{Medical Coverage}  

MediQAl covers a total of 41 medical subjects, such as cardiology, pediatrics, genetics, ophthalmology and biochemistry. The distribution of medical subjects across the dataset is displayed in Figure~\ref{fig:figure_medical_subject}. For MCQU and MCQM subsets, this information was directly available in the collected data sources. However, for the OEQ subset, the medical subjects were not consistently present across all data sources. To address this, we instructed \texttt{gpt-4o-2024-08-06}~\cite{openai2024gpt4ocard} to automatically assign a medical subject to questions when it was missing. The prompt used for this annotation is provided in Appendix~\ref{sec:appendix_prompts_med_spec}.

\section{Dataset Construction}

\subsection{Data Collection}

For the construction of this dataset, the raw data was collected from publicly available websites and forums where professors and students share examination questions intended for training purposes in preparation for the national French medical examination, such as ECN exams.

The National Classifying Tests (\textit{Épreuves Classantes Nationales - ECN}) are the theoretical exams conducted during the sixth year of medical studies in France. These exams determine the ranking of medical students, which in turn allows them to select their university hospital for residency, as well as their specialization track and the services where they will complete six-month clinical internships.

The ECN also serves as a comprehensive evaluation of the students' medical knowledge and clinical reasoning, crucial for their future roles as medical practitioners. The exam consists of the following components: (i) clinical scenario-based questions, (ii) isolated knowledge-based questions, and (iii) critical article analysis questions. The question formats include multiple-choice questions (MCQs) with five options (either a single correct answer or multiple correct answers) and open-ended short-answer questions. Each year, examination questions and answer are manually created and verified by a scientific advisory board composed of tenured academic and hospital faculty members.

In line with this structure, we organized the dataset into multiple subsets corresponding to the ECN's questions formats.
The multiple-choice questions (MCQU and MCQM) were automatically extracted from the \href{https://qcmlab.net/revision.php}{qcmlab} website in March 2024. Each instance in these subsets contains a unique ID, an optional clinical scenario, a question, five candidates, the associated medical subject and the correct answer. 
For the open-ended questions with short-answer (OEQ subset), the raw data was collected from multiple sources as HTML and PDF files. HTML files were well-structured enough to automatically extract QA instances using regular expressions. For the remaining PDF files, their structure was not homogeneous and could not be parsed automatically, and each instance was then extracted and curated manually.

\subsection{Data Filtering}

To ensure that the collected instances were homogeneous and that the questions were answerable, we applied several filters and preprocessing steps.

For the MCQU and MCQM subsets, questions with missing correct answers or candidate options were removed. Since these subsets contained a large number of questions, we used three models ({Llama-3.1-8B-Instruct~\cite{grattafiori2024llama3herdmodels}}, {Qwen2.5-7B-Instruct~\cite{qwen2.5}} and {Mistral-7B-Instruct-v0.3~\cite{jiang2023mistral7b}}) to vote on and filter questions to only keep challenging questions in the test sets. 
If any of the models answer a question correctly, the question is deemed too simple and is removed from the test sets. All removed questions were then randomly split into training (80\%) and validation (20\%) sets for both MCQU and MCQM subsets. 
The dataset splits for all tasks are presented in Table~\ref{tab:split_mediqal}.

\begin{table}[!htb]
\centering
\begin{tabular}{|c|c|c|c|}
\hline
     & Train & Validation & Test \\ \hline
MCQU & 10,113  & 2,561       & 4,343 \\ \hline
MCQM & 5,767  & 1,466       & 3,384 \\ \hline
OEQ & -     & -          & 4,969 \\ \hline
\end{tabular}
\caption{MediQAl dataset distribution}
\label{tab:split_mediqal}
\end{table}

For the OEQ subset, questions with clinical scenario containing images or tables were removed. Points awarded for each response element, sometimes embedded in the response text, were removed. Duplicates questions were identified by calculating cosine similarity on TF-IDF~\cite{sparck1972statistical} vectorized representations of both questions and answers. All QA pairs with a similarity score greater than 0.70 were manually reviewed, and duplicates were removed. To retain only short-answer questions, we tokenized each answer using a French medical tokenizer from DrBERT model~\cite{labrak-etal-2023-drbert} and excluded instances where the length of the answer exceeded 200 tokens.

\subsection{Understanding and Reasoning Questions} 

To assess the capacity of LLMs to handle complex clinical reasoning tasks beyond simple recall of medical knowledge, we implemented an automatic question categorization approach. Specifically, we categorized each question into one of two types: \textit{Understanding} or \textit{Reasoning}. This categorization was performed using \texttt{gpt-4o-2024-08-06}, following the strategy outlined by~\citet{zuo2025medxpertqabenchmarkingexpertlevelmedical}. The details of the prompt used for this process are provided in Appendix~\ref{sec:appendix_understanding_reasoning}. 

The quality of this automatic categorization was manually assessed by reviewing 10 randomly selected questions for each medical subject (5 labeled as \textit{Understanding}, and 5 as \textit{Reasoning}) from the test set of each task. In total, 858 questions were reviewed. Among these, 72 questions were explicitly mislabeled, resulting in an error rate of 8.4\%.

\section{Experiments}

\subsection{Models}

We evaluate several leading LLMs on MediQAl, covering both proprietary and open-source models, including vanilla models and recent reasoning-based models.

\paragraph{Vanilla Large Language Models:} GPT-4o-2024-08-06~\cite{openai2024gpt4ocard}, DeepSeek-V3~\cite{deepseekai2024deepseekv3technicalreport}, Qwen2.5-72B-Instruct~\cite{qwen2.5}, Llama-3.3-70B-Instruct~\cite{grattafiori2024llama3herdmodels}, Llama-3-UltraMedical 70B and 8B~\cite{zhang2024ultramedical} and BioMistral-7B~\cite{labrak-etal-2024-biomistral}.

\paragraph{Reasoning Large Language Models:} o3-2025-04-16~\cite{openai2024openaio1card}, DeepSeek-R1~\cite{deepseekai2025deepseekr1incentivizingreasoningcapability}, DeepSeek-R1-Distill-Llama-70B~\cite{deepseekai2025deepseekr1incentivizingreasoningcapability}, HuatuoGPT-o1-8B~\cite{chen2024huatuogpto1medicalcomplexreasoning}, FineMedLM-o1 8B~\cite{yu2025finemedlmo1enhancingmedicalreasoning}, DeepSeek-R1-Distill-Llama-8B~\cite{deepseekai2025deepseekr1incentivizingreasoningcapability}, DeepSeek-R1-Distill-Qwen2.5-7B~\cite{deepseekai2025deepseekr1incentivizingreasoningcapability}.

\subsection{Supervised Fine-tuning}

In addition to all evaluated models, we conducted supervised fine-tuning (SFT) on \texttt{BioMistral-7B} to assess the learnability and utility of the MediQAl dataset. The \texttt{BioMistral-7B-SFT} model was trained for two epochs using the combined training sets of all tasks (MCQU, MCQM, and OEQ).
Since the OEQ subset lacks a dedicated training set, we converted questions from MQCU and MCQM training sets into OEQ format to enable unified training. We performed full fine-tuning of the model with a learning rate of $2 \times 10^{-5}$.

\begin{table*}[!htb]
\scriptsize
\setlength\tabcolsep{10pt}
\centering
\begin{tabular}{lcccccc}
\hline

& \multicolumn{3}{c}{\textbf{EMR ($\uparrow$)}} & \multicolumn{3}{c}{\textbf{Hamming ($\uparrow$)}}  \\ \cline{2-7} 

\textbf{Model} & \textbf{Understanding} & \textbf{Reasoning} & \textbf{Avg} & \textbf{Understanding} & \textbf{Reasoning} & \textbf{Avg}   \\ \hline

\multicolumn{7}{c}{\textit{Reasoning LLMs}}   \\ 

o3 & \textbf{56.87} & \textbf{51.04} & \textbf{55.05} & \textbf{80.88} & \textbf{77.08} & \textbf{79.7} \\
DeepSeek-R1 & \underline{51.12} & \underline{43.93} & \underline{48.88} & \underline{79.21} & \underline{73.83} & \underline{77.54} \\

\hdashline

DeepSeek-R1-Distill-Llama-70B & 19.91 & 22.68 & 20.77 & 35.55 & 43.62 & 38.07 \\

\hdashline

HuatuoGPT-o1-8B \includegraphics[height=1em]{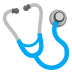} & 8.28 & 5.22 & 7.33 & 47.35 & 41.42 & 45.5 \\
FineMedLM-o1 (8B) \includegraphics[height=1em]{emoji.png} & 1.46 & 0.19 & 1.06 & 18.76 & 6.93 & 15.07 \\
DeepSeek-R1-Distill-Llama-8B & 2.1 & 2.56 & 2.25 & 10.37 & 16.06 & 12.15 \\
DeepSeek-R1-Distill-Qwen2.5-7B & 2.27 & 2.28 & 2.28 & 19.03 & 19.48 & 19.17 \\ \hline

\multicolumn{7}{c}{\textit{Vanilla LLMs}}                \\ 

GPT-4o & 46.48 & 37.37 & 43.65 & 76.03 & 69.64 & 74.04 \\
DeepSeek-V3 & 49.18 & 39.37 & 46.13 & 78.45 & 72.02 & 76.45 \\

\hdashline

Qwen2.5-72B-Instruct & 31.8 & 26.47 & 30.14 & 67.38 & 61.5 & 65.55 \\
Llama3.3-70B-Instruct & 21.72 & 11.29 & 18.47 & 62.94 & 54.41 & 60.29 \\
Llama-3-70B-UltraMedical \includegraphics[height=1em]{emoji.png} & 22.40 & 12.71 & 19.39 & 62.38 & 53.36 & 59.57 \\

\hdashline

BioMistral-7B \includegraphics[height=1em]{emoji.png} & 0.82   & 1.33  & 0.98   & 5.33   & 12.41  & 7.54 \\
Llama-3.1-8B-UltraMedical \includegraphics[height=1em]{emoji.png} & 5.11   & 4.55  & 4.93   & 44.53   & 40.39  & 43.24 \\

\hdashline

BioMistral-7B-SFT \includegraphics[height=1em]{emoji.png} & 3.21   & 3.81  & 3.38   & 24.07   & 22.75  & 23.66 \\

\hline
\end{tabular}
\caption{Performance of LLMs on the MediQAl-MCQM subset. The scores, obtained in zero-shot, are measured in terms of Exact Match Ratio (EMR) and Hamming score.}
\label{tab:MCQM}
\end{table*}

\subsection{Evaluation Framework}

All models across all tasks were evaluated in a zero-shot prompting setup, using greedy decoding for output generation when available to ensure result stability. For reasoning models that require specific evaluation settings, we followed the recommended instructions provided for each. To reduce inference time, open-source models were limited in their output length: up to 2,048 tokens for vanilla models and up to 8,192 tokens for reasoning-based models.
For API-based models (\texttt{o3}, \texttt{GPT-4o}, \texttt{DeepSeek-V3} and \texttt{DeepSeek-R1}), we followed the recommended prompting guidelines, removing the system prompt while keeping all other parameters at their default settings. 
The evaluation prompts and scripts used to extract responses from the generated text were inspired by the format of the \textit{simple-evals} framework\footnote{\href{https://github.com/openai/simple-evals}{https://github.com/openai/simple-evals}}. The specific prompts used for MCQU, MCQM and OEQ tasks are provided in Appendix~\ref{sec:appendix_prompt_mcqu}, \ref{sec:appendix_prompt_mcqm}, \ref{sec:appendix_prompt_oeq}, respectively.

\subsection{Metrics}

The evaluation metrics for each task are described below:

\paragraph{Multiple-Choice Question with Unique answer (MCQU)} For the evaluation on the MCQU subset, we used Accuracy, similarly to other single-answer multiple-choice tasks such as MMLU~\cite{hendrycks2021measuring}.

\paragraph{Multiple-Choice Question with Multiple answers (MCQM)} We used Exact Match Ratio (EMR) and Hamming score to evaluate multiple-choice questions with multiple answers, following previous work on this task~\cite{labrak-etal-2022-frenchmedmcqa}. The two metrics are defined as follows: 
\[
\text{Exact Match Ratio (EMR)} = \frac{1}{N}\sum_{i=1}^{N} [\hat{y}_i = y_i]
\]
where $N$ denotes the number of questions, $\hat{y}_i$ is the set of predicted answers for the $i^{th}$ question, $y_i$ is the set of correct answers for the $i^{th}$ question, and $[x]$ is an indicator function that returns 1 if $x$ is true and 0 otherwise.
\[
\text{Hamming Score} = \frac{1}{N} \sum_{i=1}^{N} \frac{|y_i \cap \hat{y}_i|}{|y_i \cup \hat{y}_i|}
\]
where $N$ denotes the number of questions, $y_i$ is the set of correct answers for the $i^{th}$ question, $\hat{y}_i$ is the set of predicted answers for the $i^{th}$ question, $|y_i \cap \hat{y}_i|$ is the intersection size between the correct and predicted answers, and $|y_i \cup \hat{y}_i|$ is the size of the union of correct and predicted answers.

\paragraph{Open-Ended Question with Short-Answer (OEQ)} To evaluate the free-text responses in the OEQ subset, we opted for a combination of lexical and contextual embedding-based metrics that align with human judgments on clinical texts~\cite{ben-abacha-etal-2023-investigation}. These metrics are: ROUGE-1~\cite{lin-2004-rouge}, BLEU-4~\cite{10.3115/1073083.1073135}, and BERTScore \textit{(roberta-large-mnli)}~\cite{Zhang*2020BERTScore:}.

Given the inherent complexity of evaluating open-ended question answering task, where clinically acceptable responses can differ significantly in phrasing, we supplemented traditional metrics with an automatic evaluation using a LLM-as-Judge approach~\cite{gu2025surveyllmasajudge}. This strategy consists of comparing model-generated responses with expert-provided references, allowing for more nuanced assessment beyond surface-level lexical and semantic similarity. We adopted Gemini-2.0-Flash~\cite{google_gemini2_flash_2024} as the judging model. 
The model was prompted to assign a score from 1 to 10 for each (question, model answer, expert answer) triplet, as described in the LLM-as-Judge prompt in Appendix~\ref{sec:appendix_llm_judge}.
A score of 0 was assigned to cases where the evaluated model either failed to produce a final answer or generated a response that did not conform to the expected format and was therefore unparseable. Final scores were averaged across all examples and scaled to a 0-100 range for reporting.

\section{Results and Discussion}

\paragraph{Multiple-Choice Question with Unique answer}
Table~\ref{tab:MCQU_res} shows the performance of all evaluated LLMs on the MCQU subset of the MediQAl dataset. We observe that \texttt{o3} achieves the highest performance on both \textit{Understanding} and \textit{Reasoning} questions with 73.15\% accuracy. 
Among open-source models, \texttt{DeepSeek-R1} and \texttt{DeepSeek-V3} perform well on this subset with 67.03\% and 63.32\% accuracy, even surpassing some commercial models such as \texttt{GPT-4o} (60.95\%). In contrast, models like \texttt{DeepSeek-R1-Distill-Llama-70B}, \texttt{Llama-3.3-70B} and \texttt{Qwen2.5-72B} demonstrate lower performance, correctly answering half of the questions in the test set.   
For smaller open-source models, \texttt{HuatuoGPT-o1-8B} shows impressive results compared to others in the same size category, achieving 23.49\% accuracy. Furthermore, \texttt{BioMistral-7B-SFT}, fine-tuned on the MediQAl training sets, shows substantial performance gains of 15.64\% accuracy over its base model, \texttt{BioMistral-7B}.
However, open-source reasoning-based models encounter difficulties due to token limitations during generation. Manual inspection of the generated text revealed that these models were still in the reasoning process after generating 8,192 tokens, resulting in incomplete answers which negatively impacts their performance.

\begin{table}[!htb]
\centering
\resizebox{\columnwidth}{!}{%
\begin{tabular}{lccc}
\hline
& \multicolumn{3}{c}{\textbf{Accuracy ($\uparrow$)}} \\ \cline{2-4} 
\textbf{Model}         & \textbf{Understanding} & \textbf{Reasoning} & \textbf{Avg}  \\ \hline

\multicolumn{4}{c}{\textit{Reasoning LLMs}}                \\

o3 & \textbf{74.76} & \textbf{70.63} & \textbf{73.15} \\ 
DeepSeek-R1 & \underline{69.07} & \underline{63.82} & \underline{67.03} \\

\hdashline

DeepSeek-R1-Distill-Llama-70B & 46.04 & 49.2 & 47.27 \\ 

\hdashline

HuatuoGPT-o1-8B \includegraphics[height=1em]{emoji.png} & 24.4 & 22.62 & 23.49 \\ 
FineMedLM-o1 (8B) \includegraphics[height=1em]{emoji.png} & 3.99 & 3.55 & 3.82 \\ 
DeepSeek-R1-Distill-Llama-8B & 9.31 & 6.63 & 8.27 \\ 
DeepSeek-R1-Distill-Qwen2.5-7B & 14.17 & 10.24 & 12.64 \\ \hline

\multicolumn{4}{c}{\textit{Vanilla LLMs}}                \\ 

GPT-4o & 65.0 & 54.59 & 60.95 \\ 
DeepSeek-V3 & 66.24 & 58.73 & 63.32 \\ 

\hdashline

Qwen2.5-72B-Instruct & 48.53 & 41.86 & 45.94 \\ 
Llama3.3-70B-Instruct & 46.57 & 38.31 & 43.36 \\ 
Llama-3-70B-UltraMedical \includegraphics[height=1em]{emoji.png} & 41.52 & 33.27 & 38.31 \\ 

\hdashline

BioMistral-7B \includegraphics[height=1em]{emoji.png}    & 11.72      & 12.97         & 12.20           \\
Llama-3.1-8B-UltraMedical \includegraphics[height=1em]{emoji.png} & 14.66 & 10.72 & 13.12 \\  

\hdashline

BioMistral-7B-SFT \includegraphics[height=1em]{emoji.png}    & 27.81   & 27.89  & 27.84           \\\hline

\end{tabular}%
}
\caption{Performance of LLMs on the MediQAl-MCQU subset. The scores, obtained in zero-shot, are measured with Accuracy.}
\label{tab:MCQU_res}
\end{table}

\begin{table*}[!htb]
\scriptsize
\setlength\tabcolsep{4pt}
\centering
\begin{tabular}{lcccccccccccc}
\hline

& \multicolumn{3}{c}{\textbf{ROUGE-1 ($\uparrow$)}} & \multicolumn{3}{c}{\textbf{BLEU-4 ($\uparrow$)}}  & \multicolumn{3}{c}{\textbf{BERTScore ($\uparrow$)}} & \multicolumn{3}{c}{\textbf{LLM-as-Judge ($\uparrow$)}} \\ \cline{2-13} 

\textbf{Model} & \textbf{U} & \textbf{R} & \textbf{Avg} & \textbf{U} & \textbf{R} & \textbf{Avg} & \textbf{U} & \textbf{R} & \textbf{Avg} & \textbf{U} & \textbf{R} & \textbf{Avg}  \\ \hline

\multicolumn{13}{c}{\textit{Reasoning LLMs}}   \\ 

o3 & \underline{17.61} & \underline{15.6} & \underline{16.34}  & \underline{2.56} & \underline{1.5} & \underline{1.89} & \textbf{77.65} & \textbf{76.48} & \textbf{76.91} & \textbf{87.4} & \textbf{79.07} & \textbf{82.16} \\
DeepSeek-R1 & \textbf{17.8} & \textbf{15.97} & \textbf{16.65}  & \textbf{2.68} & \textbf{1.67} & \textbf{2.04} & \underline{77.63} & \underline{76.13} & \underline{76.69} & \underline{80.26} & \underline{70.78} & \underline{74.29} \\

\hdashline

DeepSeek-R1-Distill-Llama-70B & 14.41 & 12.57 & 13.26 & 1.85 & 1.43 & 1.58 & 68.64 & 65.3 & 66.54 & 73.16 & 61.3 & 65.7 \\

\hdashline

HuatuoGPT-o1-8B \includegraphics[height=1em]{emoji.png} & 8.23 & 7.94 & 8.05 & 0.64 & 0.45 & 0.52 & 67.56 & 62.98 & 64.68 & 45.32 & 36.24 & 39.61 \\
FineMedLM-o1 (8B) \includegraphics[height=1em]{emoji.png} & 9.55 & 9.72 & 9.66 & 0.94 & 0.64 & 0.75 & 70.88 & 70.61 & 70.71 & 29.05 & 24.33 & 26.08 \\
DeepSeek-R1-Distill-Llama-8B & 5.76 & 5.41 & 5.54 & 0.56 & 0.35 & 0.43 & 60.17 & 55.75 & 57.39  & 21.24 & 16.12 & 18.02 \\
DeepSeek-R1-Distill-Qwen2.5-7B & 4.75  & 5.23  & 5.05  & 0.45  &  0.34 & 0.38 & 52.0 & 46.55 & 48.57  & 11.7 & 9.27 & 10.17 \\ \hline

\multicolumn{13}{c}{\textit{Vanilla LLMs}}                \\ 

GPT-4o & 16.29 & 14.53 & 15.18 & 2.41 & 1.29 & 1.71 & 76.47 & 75.55 & 75.89  & 77.43 & 63.77 & 68.83 \\
DeepSeek-V3 & 15.24 & 15.39 & 15.33 & 2.06 & 1.36 & 1.62 & 74.95 & 74.86 & 74.89  & 60.37 & 47.37 & 52.19 \\

\hdashline

Qwen2.5-72B-Instruct & 14.65 & 13.26 & 13.77 & 2.14 & 1.11 & 1.49 & 75.33 & 74.2 & 74.62  & 66.87 & 55.15 & 59.49 \\
Llama3.3-70B-Instruct & 14.53 & 13.59 & 13.94 & 1.7 & 1.02 & 1.27 & 74.56 & 73.49 & 73.89  & 53.32 & 43.77 & 47.31 \\

\hdashline

BioMistral-7B \includegraphics[height=1em]{emoji.png} & 6.53   & 8.61  & 7.84   & 0.67   & 0.61  & 0.63 & 44.34 & 53.95 & 50.39  & 13.64 & 13.23 & 13.38 \\
Llama-3.1-8B-UltraMedical \includegraphics[height=1em]{emoji.png} & 4.04   & 3.91  & 3.96   & 0.44   & 0.26  & 0.33  & 69.28 & 67.03 & 67.87  & 27.7 & 20.96 & 23.46 \\

\hdashline

BioMistral-7B-SFT \includegraphics[height=1em]{emoji.png} & 5.69   & 5.59  & 5.63   & 0.6   & 0.38  & 0.47 & 73.75 & 73.53 & 73.61  & 23.86 & 24.38 & 24.19 \\

\hline
\end{tabular}
\caption{Performance of LLMs on the MediQAl-OEQ subset. The scores, obtained in zero-shot, are measured with ROUGE-1, BLEU-4, BERTScore and LLM-as-Judge.}
\label{tab:OEQ}
\end{table*}

\paragraph{Multiple-Choice Question with Multiple answers}
Table~\ref{tab:MCQM} shows the performance of all evaluated LLMs on the MCQM subset of the MediQAl dataset.
The best results are achieved by \texttt{o3} with 55.05 EMR and 79.7 Hamming, followed closely by \texttt{DeepSeek-R1} (48.88 / 77.54). Vanilla models such as \texttt{DeepSeek-V3} and \texttt{GPT-4o} trail by 3-5 EMR points, indicating that additional reasoning supervision yields substantial gains. Distilled checkpoints of \texttt{DeepSeek-R1} show significant performance drops (e.g. $-$28 EMR for \texttt{DeepSeek-R1-Distill-Llama-70B}, and $-$46 EMR for \texttt{DeepSeek-R1-Distill-Llama-8B}), highlighting the trade-off imposed by aggressive model compression.
In this task, open-source reasoning-based models also face the issue of still being in the reasoning phase after generating 8,192 tokens, which negatively impacts their performance.

\paragraph{Open-Ended Question with Short-Answer}
Table~\ref{tab:OEQ} shows the performance of all evaluated LLMs on the OEQ subset of the MediQAl dataset.
For free-text answers, the performance gap widens: \texttt{o3} achieves 82.16 on the LLM-as-Judge metric, versus 74.29 for \texttt{DeepSeek-R1} and 68.83 for \texttt{GPT-4o}. 

Overlap metrics (ROUGE, BLEU and BERTScore) tend to compress differences and often yield trends that diverge from those observed with the LLM-as-Judge metric. For example, \texttt{DeepSeek-R1} outperforms \texttt{o3} on ROUGE and BLEU scores, while the opposite is observed with the LLM-as-Judge metric.

We also observed that distilled reasoning models from the DeepSeek series often reformulate the question as a multiple-choice question (MCQ) during their reasoning process, creating candidate options. This behavior poses a challenge when parsing the generated text to extract the final answer. Instead of providing a free-text response, the model tends to return the letter of one of the candidate options it created during reasoning, without necessarily including the corresponding text. This phenomenon may partly explain the comparable performance of reasoning models to vanilla models, despite their reasoning capabilities.

\begin{figure*}[t]
\centering
\subfloat[Multiple-Choice question with unique answer (MCQU)]{\label{4figs-a} \includegraphics[width=0.5\textwidth]{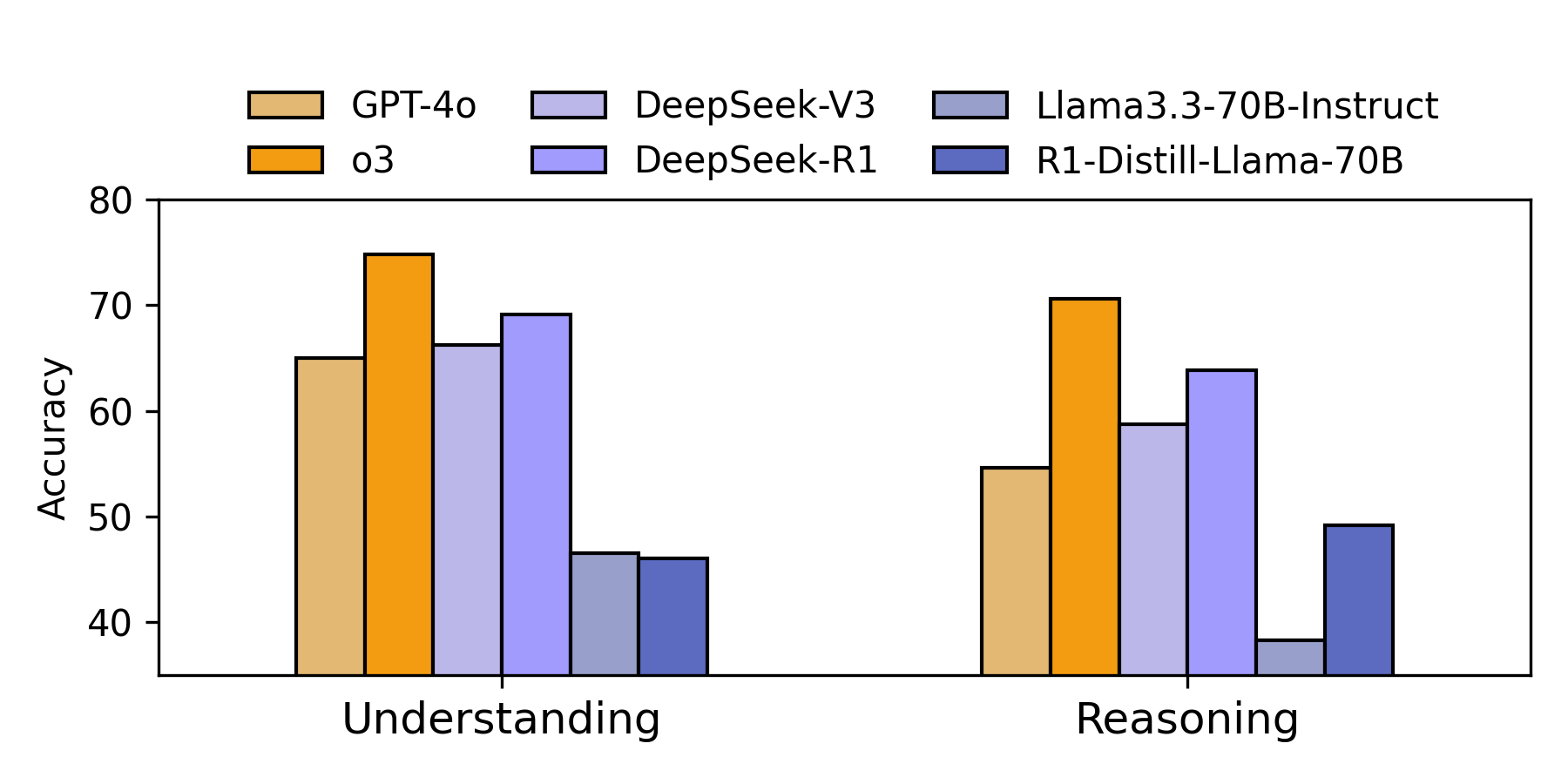}}%
\subfloat[Multiple-Choice question with multiple answers (MCQM)]{\label{4figs-b} \includegraphics[width=0.5\textwidth]{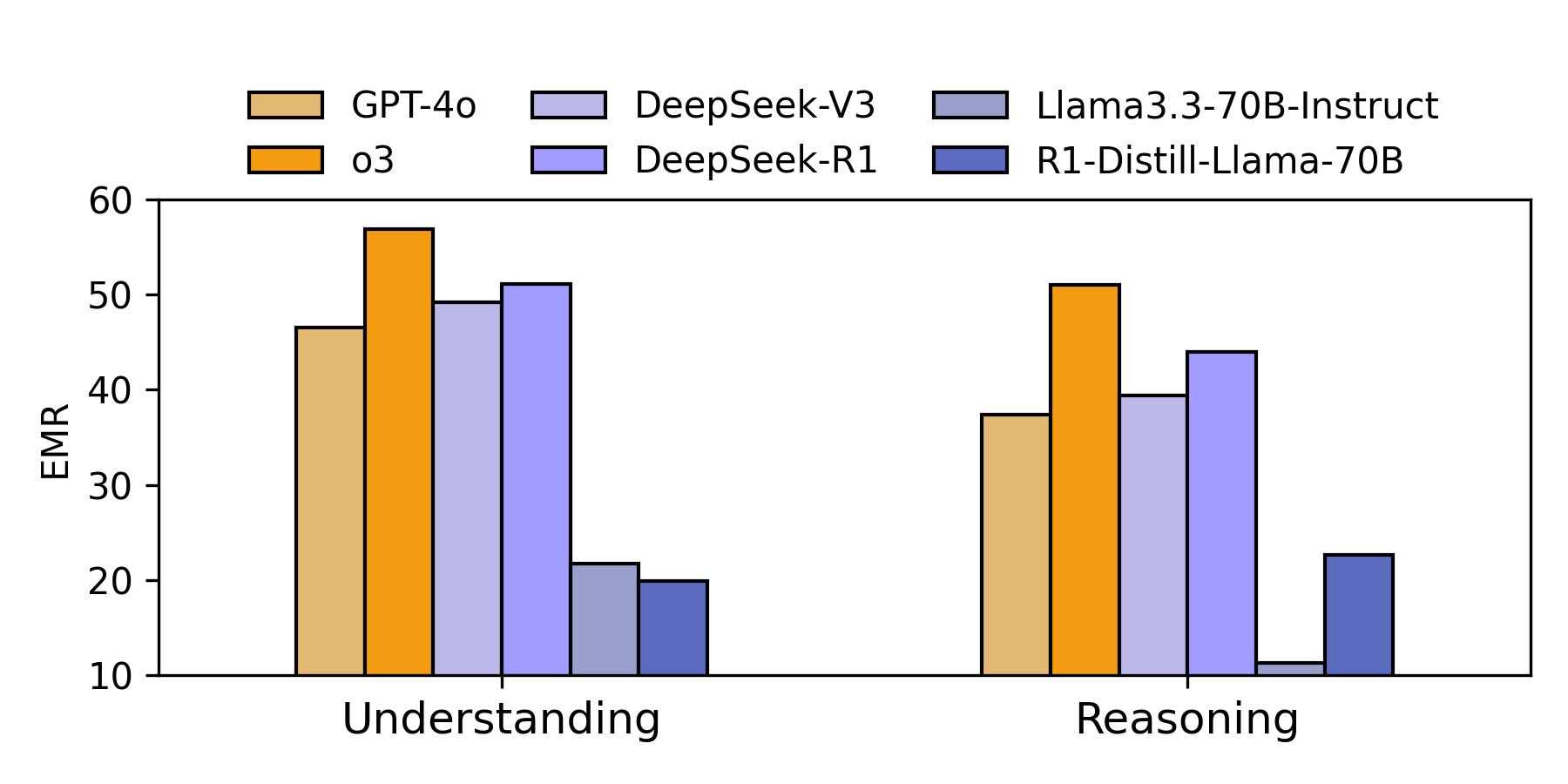}}%
\\
\subfloat[Open-ended question with short answer (OEQ)]{\label{4figs-c} \includegraphics[width=0.5\textwidth]{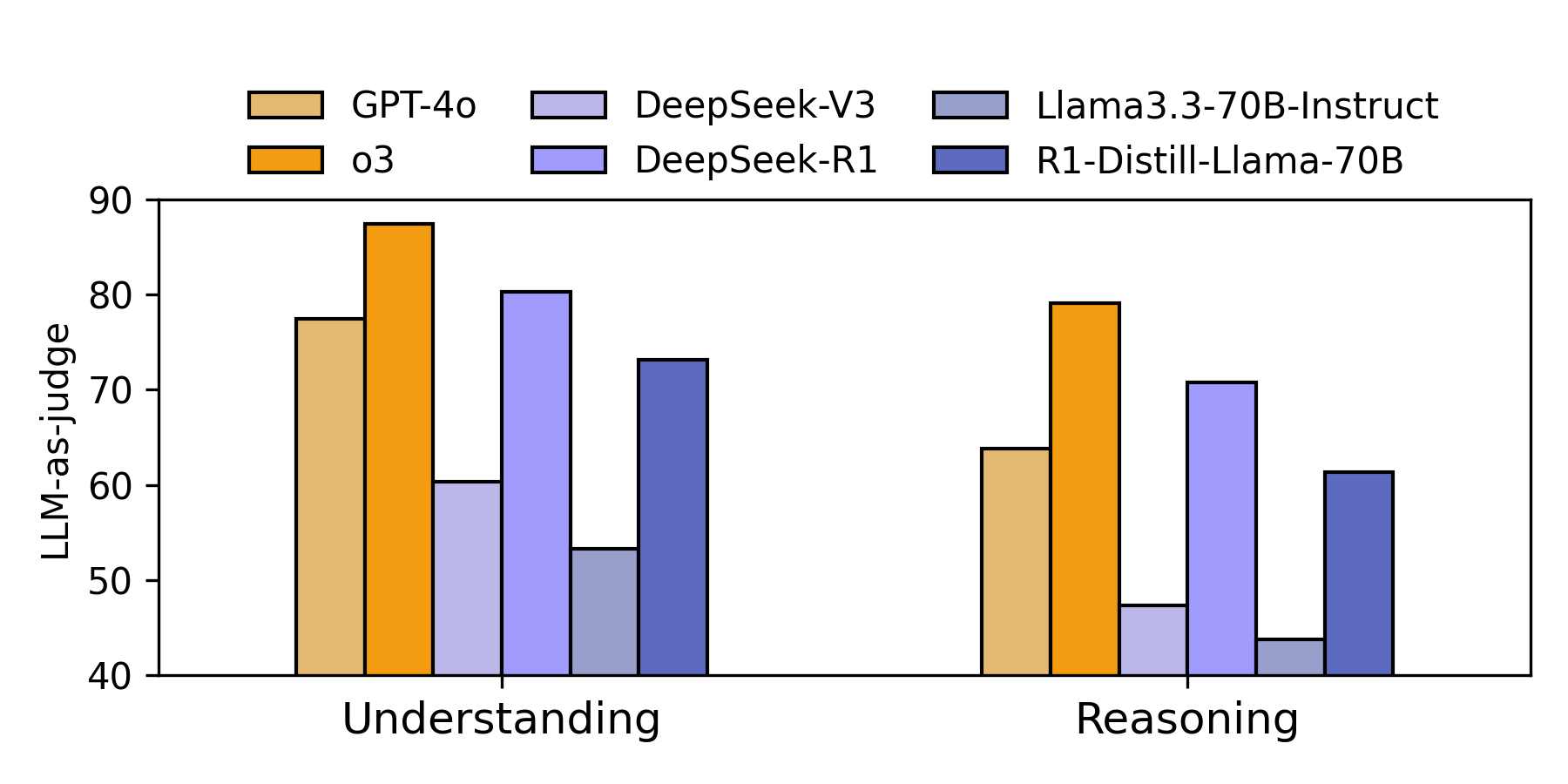}}%
\caption{Performance of three groups of models (OpenAI, DeepSeek and LLama) on all subsets of MediQAl.}
\label{4figs}
\end{figure*}

\paragraph{Medical Reasoning Performance}

Across all QA tasks, we observe a consistent performance gap between questions that require multi-step reasoning and those assessing factual recall or medical understanding. 
Averaged over all model-task combinations reported in Tables~\ref{tab:MCQU_res},~\ref{tab:MCQM} and ~\ref{tab:OEQ}, accuracy on reasoning question is 5.12 points lower than understanding questions. 
The performance gap varies across tasks: it is largest on OEQ (7.54 points), and similar for MCQU (3.90) and MCQM (3.93).
Reasoning-based models mitigate this gap to some extent but do not eliminate it. 
On MCQU and MCQM, the average performance gap for reasoning-augmented models is 2.15 and 2.02 points respectively, compared to 5.49 and 5.55 for vanilla models. In contrast, the OEQ task shows a large gap for both model types: 7.79 for vanilla and 7.29 for reasoning models.
To illustrate, on the OEQ task, \texttt{GPT-4o} show a performance gap of 13.66 points between understanding and reasoning questions, which is reduced to 8.33 with its reasoning-enhanced variant, \texttt{o3}. A similar trend is observed for the DeepSeek family: \texttt{DeepSeek-V3} shows a gap of 13.0, whereas \texttt{DeepSeek-R1} narrows this to 9.48.

Figure~\ref{4figs} presents paired performance comparisons across three model families on both \textit{Understanding} and \textit{Reasoning} questions:
\begin{itemize}
    \item \texttt{o3} vs. \texttt{GPT-4o}
    \item \texttt{DeepSeek-R1} vs. \texttt{DeepSeek-V3}
    \item \texttt{DeepSeek-R1-Distill-Llama-70B} vs. \texttt{Llama-3.3-70B-Instruct}
\end{itemize}

Two consistent trends emerge across comparisons: (i) every model performs better on \textit{Understanding} than on \textit{Reasoning} questions, except for \texttt{DeepSeek-R1-Distill-Llama-70B} on MCQU and MQCM subsets, and (ii) when comparing each reasoning model to its base version, the performance improvement is larger on reasoning questions than on understanding questions.
Theses difference underscore the impact of inference-time reasoning techniques. 
On MCQU, the average performance gain for reasoning question across the three model families is 10.67, compared to 4.02 for understanding. Similar trends are observed in MCQM, with gains of 9.87 on reasoning versus 3.51 on understanding.
For the OEQ task, performance improvements are substantial in both categories with 16.57 for understanding and 18.75 for reasoning questions.

These findings suggest that inference-time techniques, even without access to domain-specific adaptation, can significantly enhance complex medical reasoning. Nonetheless, even state-of-the-art LLMs remain well below human-level clinical reasoning in zero-shot settings. For downstream applications in healthcare, these models will require external verification or human oversight.

\paragraph{Medical Subjects Performance}

To better understand the strengths and weaknesses of LLMs on our dataset, we analyzed their performance across individual medical subjects for each QA task (see Tables~\ref{tab:MCQU-subjects},~\ref{tab:MCQM-subjects} and~\ref{tab:OEQ-subjects} in Appendix). In the MCQU task, the models performed best on subjects such as genetics, anatomy, dermatology, physiology, otorhinolaryngology (ENT), ophtalmology, neurology and hematology, all achieving over 80\% accuracy. Conversely, subjects like cytology (notably low at 16.67\% due to limited limited examples in the dataset), epidemiology, and psychiatry showed the lowest performance with accuracy below 60\%.
In the MQCM task, the easiest subjects for LLMs were dermatology, genetics, and microbiology (all above 65\% EMR), while rehabilitation, occupational medicine, and pathological anatomy were the most challenging, with scores under 40\% EMR.
Finally, in the OEQ task, the best-performing subjects were bacteriology, parasitology, and semiology, each with LLM-as-Judge scores above 90\%, whereas occupational medicine and endocrinology and metabolism were among the lowest, with score falling below 70\%. These results highlight that LLMs' capabilities vary significantly by medical domain and question type, with certain specialized or interdisciplinary fields remaining particularly challenging.

\section{Conclusion}

In this work, we introduce MediQAl, a novel dataset for medical question answering in French. This dataset includes three tasks: (i) Multiple-Choice Question with Unique answer (MCQU), (ii) Multiple-Choice Question with Multiple answers (MCQM), and (iii) Open-Ended Question with Short-Answer (OEQ). MediQAl covers a wide range of medical subjects and is designed to challenge models' reasoning and comprehension across various cognitive tasks. 

This work addresses significant gaps in current medical benchmarks by introducing new tasks for French and expanding resources beyond English and Chinese. We evaluated 14 models on MediQAl and demonstrated that reasoning-based models outperform vanilla LLMs on various question answering tasks.

\section*{Acknowledgments}

This work was performed using HPC resources from GENCI-IDRIS (Grant 2024-
AD011013715R2). This work was financially supported by ANR
MALADES (ANR-23-IAS1-0005).

\section*{Ethical Statement}

This study required substantial computational resources, for a total of approximately 4,000 hours on A100 80GB GPUs. These resources were dedicated to models evaluations, experimentation with various models, and debugging. According to documentation from the Jean Zay supercomputer\footnote{\href{http://www.idris.fr/media/jean-zay/jean-zay-conso-heure-calcul.pdf}{http://www.idris.fr/media/jean-zay/jean-zay-conso-heure-calcul.pdf}}, the total environmental cost amounted to 1,036,000 Wh or 59.05 kg CO2eq, based on the carbon intensity of the energy grid as reported in the BLOOM environmental cost study conducted on Jean Zay~\cite{luccioni2022estimating}. Additionally, the total inference cost on API for the data augmentation strategies, the LLM-as-Judge evaluation, and the zero-shot evaluation of o3, GPT-4o, DeepSeek-R1 and DeepSeek-V3 on the three tasks amounted to 346 USD.

\bibliography{custom}

\onecolumn

\appendix

\section{Prompts}
\label{sec:appendix}

\subsection{Medical Subjects Prompt}
\label{sec:appendix_prompts_med_spec}

\begin{figure}[htb]
\centering
\begin{instructionframe}{Medical Subjects Annotation Prompt}
\textit{You are an experienced medical doctor and independent practitioner.}
\textit{Your task will be to label a clinical scenario according to the medical subject it corresponds to.}

\vspace{0.5em}

\textit{You will be given a list of medical subjects, followed by a clinical scenario. Please determine which subject the clinical scenario best pertains to. If the clinical scenario is related to multiple subjects, only select the most relevant one.}

\vspace{0.5em}

\textit{Directly output the name of the final subject you selected from the list of available subjects.}

\vspace{0.5em} 

**\textbf{Subjects:}**

\vspace{0.5em}

Cardiologie et Pathologie Vasculaire, Hépato-Gastro-Entérologie, Pneumologie, Néphro-Urologie, Psychiatrie, Hématologie, Endocrinologie-Métabolisme, Gynécologie-Obstétrique, Rhumatologie, Neurologie, Maladies Infectieuses, Dermatologie, Pédiatrie, Oto-Rhino-Laryngologie, Ophtalmologie, Immunologie, Orthopédie, Pharmacologie, Médecine Légale et Toxicologie, Anatomie pathologique, Biochimie, Epidémiologie, Génétique, Médecine du Travail, Microbiologie, Parasitologie, Rééducation, Physiologie, Histologie, Radiologie, Cytologie, Embryologie, Anatomie, Urgences

\vspace{0.5em}

**\textbf{Clinical scenario:}** 

\vspace{0.5em}

\colorbox{black!5}{\{clinical\_scenario\}}

\vspace{0.5em}

**\textbf{Output:}**
\end{instructionframe}
\caption{Prompt for Medical Subjects Annotation in the MediQAl-OEQ subset. The list of medical subjects in the prompt was made from the list of medical subjects from MCQU and MCQM subsets.}
\label{prompt:medical_subject}
\end{figure}

\newpage

\subsection{Understanding or Reasoning Annotation Prompt}
\label{sec:appendix_understanding_reasoning}

\begin{figure}[htb]
\centering
\begin{instructionframe}{Understanding/Reasoning Annotation Prompt}
\textit{You are an experienced medical doctor and independent practitioner.}
\textit{Your task will be to determine whether a medical question primarily challenges the answerer's medical knowledge understanding or medical reasoning skills.}

\vspace{0.5em}

\textit{You will be given the question. Please determine whether the question primarily challenges the answerer's medical knowledge understanding or medical reasoning ability.}

\textit{Reasoning : complicated, reasoning-heavy questions.}

\textit{Understanding : little to no reasoning and instead assess skills such as medical knowledge.}

\vspace{0.5em}

\textit{Directly output either "Understanding" Or "Reasoning" as your answer without any additional information or explanations.}

\vspace{0.5em}

**\textbf{Clinical scenario:}** 

\vspace{0.5em}

\colorbox{black!5}{\{clinical\_scenario\}}

\vspace{0.5em}

**\textbf{Question:}** 

\vspace{0.5em}

\colorbox{black!5}{\{question\}}

\vspace{0.5em}

**\textbf{Output:}**
\end{instructionframe}
\caption{Prompt for labeling questions as Reasoning or Understanding.}
\label{prompt:vanilla_mqcu}
\end{figure}

\newpage

\subsection{Prompt for Zero-Shot Evaluation on the MCQU Subset}
\label{sec:appendix_prompt_mcqu}

\begin{figure}[htb]
\centering
\begin{instructionframe}{MCQU Prompt}
\textit{You are an experienced medical doctor and independent practitioner.}
\textit{Your task is to answer the following medical multiple-choice question. There is only one correct choice. The last line of your response should be of the following format: 'Answer: \$LETTER' (without quotes) where LETTER is one of ABCDE. Think step by step before answering.}

\vspace{0.3em}

**\textbf{Clinical scenario:}** 

\vspace{0.3em}

\colorbox{black!5}{\{clinical\_scenario\}}

\vspace{0.3em}

**\textbf{Question:}** 

\vspace{0.3em}

\colorbox{black!5}{\{question\}}

\vspace{0.3em}

\textbf{(A)} \colorbox{black!5}{\{option\_A\}}

\textbf{(B)} \colorbox{black!5}{\{option\_B\}}

\textbf{(C)} \colorbox{black!5}{\{option\_C\}}

\textbf{(D)} \colorbox{black!5}{\{option\_D\}}

\textbf{(E)} \colorbox{black!5}{\{option\_E\}}

\end{instructionframe}
\caption{Prompt for zero-shot evaluation of LLMs on the MCQU subset. The clinical scenario is optional in the prompt.}
\label{prompt:mcqu}
\end{figure}

\newpage

\subsection{Prompt for Zero-Shot Evaluation of LLMs on the MCQM Subset}
\label{sec:appendix_prompt_mcqm}

\begin{figure}[htb]
\centering
\begin{instructionframe}{MCQM Prompt}
\textit{You are an experienced medical doctor and independent practitioner.}
\textit{Your task is to answer the following medical multiple-choice question. Multiple selections are required; single-choice answers are not accepted. The last line of your response should be of the following format: 'Answer: \$LETTERS' (without quotes) where LETTERS are multiple letters of ABCDE, separated by commas (e.g., A,B,C). Think step by step before answering.}

\vspace{0.3em}

**\textbf{Clinical scenario:}** 

\vspace{0.3em}

\colorbox{black!5}{\{clinical\_scenario\}}

\vspace{0.3em}

**\textbf{Question:}** 

\vspace{0.3em}

\colorbox{black!5}{\{question\}}

\vspace{0.3em}

\textbf{(A)} \colorbox{black!5}{\{option\_A\}}

\textbf{(B)} \colorbox{black!5}{\{option\_B\}}

\textbf{(C)} \colorbox{black!5}{\{option\_C\}}

\textbf{(D)} \colorbox{black!5}{\{option\_D\}}

\textbf{(E)} \colorbox{black!5}{\{option\_E\}}

\end{instructionframe}
\caption{Prompt for zero-shot evaluation of LLMs on the MCQM subset. The clinical scenario is optional in the prompt.}
\label{prompt:mcqm}
\end{figure}

\subsection{Prompt for Zero-Shot Evaluation of LLMs on the OEQ Subset}
\label{sec:appendix_prompt_oeq}

\begin{figure}[htb]
\centering
\begin{instructionframe}{OEQ Prompt}
\textit{You are an experienced medical doctor and independent practitioner.}
\textit{Your task is to answer the following medical question in French by providing a well-structured and concise response. The last line of your response should be of the following format: 'Answer: \$TEXT' (without quotes) where TEXT is your final answer in French. Think step by step before answering.}

\vspace{0.3em}

**\textbf{Clinical scenario:}** 

\vspace{0.3em}

\colorbox{black!5}{\{clinical\_scenario\}}

\vspace{0.3em}

**\textbf{Question:}** 

\vspace{0.3em}

\colorbox{black!5}{\{question\}}

\end{instructionframe}
\caption{Prompt for zero-shot evaluation of LLMs on the OEQ subset. The clinical scenario is optional in the prompt.}
\label{prompt:OEQ}
\end{figure}

\newpage

\subsection{Prompt for LLM-as-Judge Evaluation on the OEQ Subset}
\label{sec:appendix_llm_judge}

\begin{figure}[htb]
\centering
\begin{instructionframe}{LLM-as-Judge Prompt}
[System]

Please act as an impartial judge and evaluate the quality of the response provided by an AI assistant to the French medical question displayed below. Your evaluation should consider clinical correctness, factual coverage and the impact of differences between the answers on patient safety and care. You will be given a reference answer (Expert-provided answer) and the assistant's answer (Model-generated answer). Your job is to evaluate how closely a Model-generated answer aligns with an Expert-provided answer. Base your judgment only on the Expert's provided answer, and never rely on your own medical knowledge or external resources. Begin your evaluation by comparing the assistant's answer with the reference answer. Avoid any position biases and ensure that the order in which the responses were presented does not influence your decision. Do not allow the length of the responses to influence your evaluation. Be as objective as possible. After providing your explanation, you must rate the response on a scale of 1 to 10 by strictly following this format "Rating: [[rating]]", for example: "Rating: [[5]]".

\vspace{0.3em}

[Medical Question]

\colorbox{black!5}{\{question\}}

\vspace{0.3em}

[The Start of Expert Answer]

\colorbox{black!5}{\{answer\_ref\}}

[The End of Expert Answer]

\vspace{0.3em}

[The Start of Assistant Answer]

\colorbox{black!5}{\{answer\_a\}}
    
[The End of Assistant Answer]

\end{instructionframe}
\caption{Prompt for LLM-as-judge evaluation of LLMs on the OEQ subset.}
\label{prompt:llm_judge}
\end{figure}

\newpage

\section{Models Performance on Medical Subjects}
\label{sec:appendix_medical_subject}

\vspace{4cm}

\begin{table}[ht!]
\tiny
\begin{tabular}{l|l|l|l|l|l|l|l|l|l|l|l|l|l|l|l}
\textbf{Medical Subject}                           & \rotatebox{60}{\makebox[0pt][l]{\textbf{o3}}} & \rotatebox{60}{\makebox[0pt][l]{\textbf{DeepSeek-R1}}} & \rotatebox{60}{\makebox[0pt][l]{\textbf{DeepSeek-R1-Distill-Llama-70B}}} & \rotatebox{60}{\makebox[0pt][l]{\textbf{HuatuoGPT-o1-8B}}} & \rotatebox{60}{\makebox[0pt][l]{\textbf{FineMedLM-o1}}} & \rotatebox{60}{\makebox[0pt][l]{\textbf{DeepSeek-R1-Distill-Llama-8B}}} & \rotatebox{60}{\makebox[0pt][l]{\textbf{DeepSeek-R1-Distill-Qwen2.5-7B}}} & \rotatebox{60}{\makebox[0pt][l]{\textbf{GPT-4o}}} & \rotatebox{60}{\makebox[0pt][l]{\textbf{DeepSeek-V3}}} & \rotatebox{60}{\makebox[0pt][l]{\textbf{Qwen2.5-72B-Instruct}}} & \rotatebox{60}{\makebox[0pt][l]{\textbf{Llama3.3-70B-Instruct}}} & \rotatebox{60}{\makebox[0pt][l]{\textbf{Llama-3-70B-UltraMedical}}} & \rotatebox{60}{\makebox[0pt][l]{\textbf{BioMistral-7B}}} & \rotatebox{60}{\makebox[0pt][l]{\textbf{Llama-3.1-8B-UltraMedical}}} & \rotatebox{60}{\makebox[0pt][l]{\textbf{BioMistral-7B-SFT}}} \\ \hline
Anatomy & 77.78 & \textbf{88.89} & 33.33 & 33.33 & 0.0 & 16.67 & 33.33 & 77.78 & 83.33 & 50.0 & 38.89 & 50.0 & 22.22 & 11.11 & 22.22 \\
Pathological Anatomy &  \textbf{66.67} & 61.54 & 37.18 & 25.64 & 6.41 & 7.69 & 14.1 & 53.85 & 53.85 & 47.44 & 35.9 & 37.18 & 14.1 & 24.36 & 15.38 \\
Biochemistry &  71.84 & \textbf{73.79} & 46.6 & 27.18 & 2.91 & 14.56 & 14.56 & 65.05 & 65.05 & 52.43 & 43.69 & 33.01 & 14.56 & 12.62 & 30.1 \\
Cardiology & \textbf{78.59} & 74.31 & 52.6 & 26.61 & 2.45 & 6.42 & 11.01 & 66.67 & 67.89 & 46.48 & 46.48 & 37.61 & 12.23 & 12.84 & 27.83 \\
Cytology & 16.67 & 16.67 & 16.67 & 0.0 & 0.0 & 0.0 & \textbf{50.0} & 16.67 & 16.67 & 16.67 & 16.67 & 16.67 & 16.67 & 0.0 & 16.67 \\
Dermatology &  \textbf{83.2} & 73.6 & 50.4 & 16.8 & 4.8 & 5.6 & 14.4 & 70.4 & 76.0 & 51.2 & 52.0 & 38.4 & 14.4 & 13.6 & 41.6 \\
Embryology &  \textbf{66.67} & \textbf{66.67} & 33.33 & 33.33 & 0.0 & 0.0 & 33.33 & \textbf{66.67} & \textbf{66.67} & \textbf{66.67} & \textbf{66.67} & 0.0 & 33.33 & 33.33 & \textbf{66.67} \\
Endocrinology and Metabolism & \textbf{72.95} & 67.15 & 48.31 & 21.74 & 1.93 & 7.25 & 8.21 & 58.94 & 61.84 & 42.51 & 41.06 & 38.65 & 11.59 & 13.53 & 25.6 \\
Epidemiology & \textbf{53.61} & 49.48 & 39.18 & 20.62 & 4.12 & 12.37 & 22.68 & 43.3 & 45.36 & 40.21 & 34.02 & 25.77 & 10.31 & 8.25 & 19.59 \\
Gynecology and Obstetrics & \textbf{68.93} & 62.86 & 45.0 & 14.29 & 1.79 & 5.0 & 8.93 & 52.14 & 55.0 & 44.29 & 41.79 & 40.36 & 10.36 & 12.14 & 28.57 \\
Genetics & \textbf{88.89} & \textbf{88.89} & 55.56 & 25.0 & 11.11 & 12.5 & 12.5 & 73.61 & 81.94 & 62.5 & 61.11 & 48.61 & 12.5 & 25.0 & 31.94 \\
Histology & \textbf{76.06} & 60.56 & 43.66 & 16.9 & 4.23 & 14.08 & 15.49 & 63.38 & 59.15 & 46.48 & 45.07 & 30.99 & 8.45 & 16.9 & 18.31 \\
Hematology & \textbf{80.76} & 71.61 & 52.05 & 21.77 & 4.73 & 6.31 & 9.78 & 62.78 & 67.19 & 47.95 & 42.59 & 38.17 & 16.72 & 13.88 & 33.12 \\
Hepato-Gastroenterology & \textbf{64.03} & 59.71 & 41.01 & 23.38 & 4.32 & 6.12 & 12.23 & 52.88 & 56.47 & 41.01 & 40.29 & 34.17 & 12.23 & 11.51 & 24.1 \\
Immunology & 68.13 & \textbf{70.33} & 52.75 & 24.18 & 3.3 & 8.79 & 14.29 & 62.64 & 61.54 & 57.14 & 43.96 & 39.56 & 7.69 & 13.19 & 25.27 \\
Infectious Diseases & \textbf{73.96} & 66.15 & 47.92 & 22.4 & 2.6 & 8.33 & 10.94 & 62.5 & 64.58 & 49.48 & 40.62 & 37.5 & 13.02 & 7.81 & 25.52 \\
Microbiology & \textbf{68.25} & 63.49 & 36.51 & 33.33 & 3.17 & 19.05 & 15.87 & 66.67 & 58.73 & 46.03 & 47.62 & 42.86 & 12.7 & 12.7 & 26.98 \\
Forensic Medicine and Toxicology & \textbf{72.73} & 47.47 & 43.43 & 20.2 & 3.03 & 10.1 & 15.15 & 53.54 & 51.52 & 36.36 & 35.35 & 27.27 & 11.11 & 11.11 & 21.21 \\
Occupational Medicine & \textbf{77.57} & 70.09 & 41.12 & 20.56 & 2.8 & 10.28 & 13.08 & 55.14 & 62.62 & 38.32 & 42.99 & 41.12 & 9.35 & 13.08 & 28.97 \\
Neurology & \textbf{80.79} & 72.88 & 53.11 & 25.99 & 3.95 & 8.47 & 14.12 & 68.93 & 75.14 & 46.89 & 44.07 & 40.11 & 12.43 & 13.56 & 19.77 \\
Nephro-Urology & \textbf{73.57} & 71.37 & 48.46 & 30.84 & 2.64 & 5.73 & 9.69 & 64.32 & 68.28 & 48.02 & 43.61 & 41.85 & 11.89 & 11.89 & 27.75 \\
Ophthalmology & \textbf{83.16} & 80.0 & 53.68 & 28.42 & 8.42 & 11.58 & 9.47 & 80.0 & 76.84 & 63.16 & 56.84 & 62.11 & 9.47 & 22.11 & 32.63 \\
Orthopedics & \textbf{70.93} & 64.53 & 52.33 & 25.58 & 3.49 & 6.98 & 15.7 & 58.72 & 64.53 & 47.67 & 45.93 & 33.72 & 12.79 & 10.47 & 28.49 \\
Otorhinolaryngology (ENT) & \textbf{81.89} & 75.59 & 55.12 & 25.98 & 3.15 & 8.66 & 11.81 & 73.23 & 72.44 & 50.39 & 47.24 & 44.09 & 9.45 & 17.32 & 37.8 \\
Parasitology & \textbf{78.26} & 63.04 & 43.48 & 28.26 & 8.7 & 6.52 & 13.04 & 65.22 & 63.04 & 45.65 & 50.0 & 52.17 & 4.35 & 17.39 & 23.91 \\
Pharmacology & \textbf{72.22} & 66.67 & 53.7 & 35.19 & 7.41 & 16.67 & 14.81 & 62.96 & 61.11 & 55.56 & 57.41 & 37.04 & 14.81 & 12.96 & 37.04 \\
Physiology & 82.61 & 82.61 & 50.72 & 17.39 & 1.45 & 15.94 & 18.84 & 82.61 & \textbf{84.06} & 52.17 & 50.72 & 49.28 & 13.04 & 10.14 & 28.99 \\
Pulmonology & \textbf{66.13} & 58.87 & 44.76 & 21.77 & 4.84 & 8.06 & 11.29 & 54.44 & 56.85 & 42.74 & 42.34 & 36.69 & 12.9 & 12.5 & 29.44 \\
Psychiatry & \textbf{59.86} & 48.59 & 38.73 & 21.83 & 1.41 & 9.15 & 12.68 & 50.0 & 45.77 & 28.87 & 42.96 & 37.32 & 11.97 & 9.86 & 29.58 \\
Pediatrics & \textbf{69.77} & 66.51 & 47.44 & 23.26 & 6.05 & 9.77 & 14.88 & 59.07 & 60.93 & 50.23 & 39.07 & 33.49 & 12.56 & 13.49 & 30.7 \\
Radiology & \textbf{75.0} & 71.43 & 25.0 & 7.14 & 0.0 & 3.57 & 17.86 & 64.29 & 71.43 & 39.29 & 28.57 & 28.57 & 10.71 & 10.71 & 17.86 \\
Rheumatology & \textbf{74.64} & 68.42 & 45.45 & 28.23 & 4.78 & 6.22 & 13.88 & 57.42 & 63.64 & 36.84 & 37.8 & 39.23 & 11.48 & 13.88 & 24.88 \\ \hline
\end{tabular}
\caption{Performance by medical subject on the MediQAl-MCQU subset. The scores, obtained in zero-shot, are measured with Accuracy.}
\label{tab:MCQU-subjects}
\end{table}

\begin{table}[t]
\tiny
\begin{tabular}{l|l|l|l|l|l|l|l|l|l|l|l|l|l|l|l}
\textbf{Medical Subject}                           & \rotatebox{60}{\makebox[0pt][l]{\textbf{o3}}} & \rotatebox{60}{\makebox[0pt][l]{\textbf{DeepSeek-R1}}} & \rotatebox{60}{\makebox[0pt][l]{\textbf{DeepSeek-R1-Distill-Llama-70B}}} & \rotatebox{60}{\makebox[0pt][l]{\textbf{HuatuoGPT-o1-8B}}} & \rotatebox{60}{\makebox[0pt][l]{\textbf{FineMedLM-o1}}} & \rotatebox{60}{\makebox[0pt][l]{\textbf{DeepSeek-R1-Distill-Llama-8B}}} & \rotatebox{60}{\makebox[0pt][l]{\textbf{DeepSeek-R1-Distill-Qwen2.5-7B}}} & \rotatebox{60}{\makebox[0pt][l]{\textbf{GPT-4o}}} & \rotatebox{60}{\makebox[0pt][l]{\textbf{DeepSeek-V3}}} & \rotatebox{60}{\makebox[0pt][l]{\textbf{Qwen2.5-72B-Instruct}}} & \rotatebox{60}{\makebox[0pt][l]{\textbf{Llama3.3-70B-Instruct}}} & \rotatebox{60}{\makebox[0pt][l]{\textbf{Llama-3-70B-UltraMedical}}} & \rotatebox{60}{\makebox[0pt][l]{\textbf{BioMistral-7B}}} & \rotatebox{60}{\makebox[0pt][l]{\textbf{Llama-3.1-8B-UltraMedical}}} & \rotatebox{60}{\makebox[0pt][l]{\textbf{BioMistral-7B-SFT}}} \\ \hline
Pathological Anatomy  &  \textbf{43.4} & 41.51 & 9.43 & 7.55 & 1.89 & 1.89 & 1.89 & 28.3 & 32.08 & 20.75 & 16.98 & 15.09 & 0.0 & 7.55 & 1.38 \\
Biochemistry  &  46.67 & 48.89 & 22.22 & 11.11 & 0.0 & 0.0 & 2.22 & 46.67 & \textbf{51.11} & 42.22 & 20.0 & 22.22 & 0.0 & 0.0 & 0.0 \\
Cardiology  &  \textbf{62.03} & 54.43 & 27.85 & 8.44 & 1.27 & 2.11 & 2.95 & 49.79 & 48.52 & 29.96 & 18.57 & 23.21 & 0.84 & 5.49 & 5.41 \\
Dermatology  &  \textbf{66.67} & 58.49 & 27.04 & 4.4 & 0.0 & 2.52 & 4.4 & 47.8 & 54.09 & 35.22 & 17.61 & 16.35 & 1.26 & 3.77 & 3.28 \\
Endocrinology and Metabolism  &  \textbf{48.89} & 45.0 & 18.89 & 8.89 & 1.67 & 1.67 & 2.22 & 39.44 & 45.56 & 29.44 & 19.44 & 21.11 & 1.67 & 8.89 & 6.66 \\
Epidemiology  &  \textbf{46.15} & 38.46 & 21.54 & 10.77 & 1.54 & 1.54 & 10.77 & 35.38 & 44.62 & 23.08 & 15.38 & 27.69 & 1.54 & 9.23 & 2.41 \\
Gynecology and Obstetrics  &  \textbf{50.27} & 37.3 & 16.22 & 7.57 & 0.54 & 4.86 & 2.16 & 36.22 & 37.3 & 25.95 & 23.24 & 14.59 & 2.7 & 3.78 & 3.37 \\
Genetics  &  \textbf{71.64} & 67.16 & 29.85 & 13.43 & 1.49 & 2.99 & 4.48 & 61.19 & 65.67 & 41.79 & 32.84 & 29.85 & 1.49 & 7.46 & 5.65 \\
Hematology  &  \textbf{62.96} & 58.33 & 25.93 & 8.8 & 0.0 & 2.31 & 2.31 & 48.15 & 50.93 & 31.94 & 14.35 & 17.59 & 0.46 & 6.48 & 4.35 \\
Hepato-Gastroenterology  &  \textbf{50.0} & 45.65 & 16.96 & 6.09 & 1.3 & 1.3 & 3.48 & 38.26 & 42.61 & 25.65 & 16.09 & 15.22 & 0.0 & 1.74 & 1.29 \\
Immunology  &  \textbf{55.22} & 53.73 & 23.88 & 10.45 & 1.49 & 2.99 & 1.49 & 46.27 & 50.75 & 35.82 & 34.33 & 29.85 & 0.0 & 5.97 & 2.89 \\
Infectious Diseases  &  \textbf{47.14} & 41.43 & 16.43 & 5.0 & 0.71 & 5.0 & 0.71 & 35.71 & 42.14 & 24.29 & 17.86 & 15.0 & 1.43 & 4.29 & 1.95 \\
Microbiology  &  65.79 & 60.53 & 36.84 & 15.79 & 0.0 & 0.0 & 0.0 & 60.53 & \textbf{71.05} & 31.58 & 36.84 & 21.05 & 2.63 & 5.26 & 4.99 \\
Forensic Medicine and Toxicology  &  \textbf{47.62} & 27.62 & 19.05 & 2.86 & 1.9 & 2.86 & 1.9 & 35.24 & 32.38 & 24.76 & 18.1 & 13.33 & 0.0 & 2.86 & 1.75 \\
Occupational Medicine  &  \textbf{38.33} & 28.33 & 8.33 & 6.67 & 0.0 & 1.67 & 5.0 & 30.0 & 31.67 & 18.33 & 6.67 & 10.0 & 0.0 & 1.67 & 0.51 \\
Neurology  &  \textbf{63.29} & 53.16 & 22.78 & 10.13 & 1.27 & 1.9 & 1.9 & 55.06 & 52.53 & 37.97 & 20.89 & 28.48 & 0.63 & 6.33 & 4.95 \\
Nephro-Urology  &  \textbf{56.34} & 52.58 & 23.0 & 4.69 & 1.88 & 0.94 & 1.88 & 43.19 & 45.54 & 31.92 & 19.25 & 17.37 & 0.0 & 5.16 & 4.55 \\
Ophthalmology  &  \textbf{59.23} & 57.69 & 27.69 & 6.92 & 1.54 & 0.77 & 0.0 & 47.69 & 54.62 & 36.92 & 26.15 & 30.0 & 2.31 & 1.54 & 2.12 \\
Orthopedics  &  \textbf{54.37} & 41.75 & 22.33 & 3.88 & 0.97 & 1.94 & 2.91 & 40.78 & 43.69 & 31.07 & 11.65 & 13.59 & 0.97 & 4.85 & 3.11 \\
Otorhinolaryngology (ENT)  &  \textbf{62.77} & 54.01 & 14.6 & 8.76 & 1.46 & 4.38 & 1.46 & 51.82 & 46.72 & 30.66 & 19.71 & 19.71 & 2.92 & 5.11 & 4.88 \\
Parasitology  &  56.0 & \textbf{64.0} & 24.0 & 16.0 & 0.0 & 0.0 & 0.0 & 56.0 & 56.0 & 44.0 & 28.0 & 32.0 & 0.0 & 4.0 & 3.14 \\
Pharmacology  &  49.23 & 46.15 & 23.08 & 13.85 & 3.08 & 6.15 & 3.08 & 47.69 & \textbf{50.77} & 27.69 & 20.0 & 20.0 & 0.0 & 6.15 & 5.64 \\
Pulmonology  &  \textbf{45.97} & 43.13 & 14.69 & 4.27 & 0.95 & 0.95 & 0.95 & 34.12 & 39.34 & 27.49 & 14.22 & 15.64 & 1.42 & 6.64 & 6.33 \\
Psychiatry  &  \textbf{47.88} & 42.42 & 16.36 & 6.67 & 0.0 & 1.82 & 1.21 & 36.97 & 36.97 & 26.06 & 9.09 & 19.39 & 0.61 & 4.85 & 4.75 \\
Pediatrics  &  \textbf{52.41} & 47.59 & 17.93 & 5.52 & 0.69 & 2.76 & 1.38 & 42.07 & 42.07 & 27.59 & 11.03 & 16.55 & 1.38 & 4.14 & 3.88 \\
Rheumatology  &  \textbf{64.48} & 60.11 & 21.31 & 7.65 & 1.64 & 1.64 & 1.64 & 55.19 & 56.28 & 34.43 & 24.04 & 21.86 & 0.0 & 4.37 & 2.05 \\
Rehabilitation  &  0.0 & 0.0 & 0.0 & 0.0 & 0.0 & 0.0 & 0.0 & 0.0 & 0.0 & \textbf{50.0} & 0.0 & 0.0 & 0.0 & 0.0 & 0.0 \\ \hline
\end{tabular}
\caption{Performance by medical subject on the MediQAl-MCQM subset. The scores, obtained in zero-shot, are measured with EMR.}
\label{tab:MCQM-subjects}
\end{table}

\begin{table}[t]
\tiny
\begin{tabular}{l|l|l|l|l|l|l|l|l|l|l|l|l|l|l}
\textbf{Medical Subject}                           & \rotatebox{60}{\makebox[0pt][l]{\textbf{o3}}} & \rotatebox{60}{\makebox[0pt][l]{\textbf{DeepSeek-R1}}} & \rotatebox{60}{\makebox[0pt][l]{\textbf{DeepSeek-R1-Distill-Llama-70B}}} & \rotatebox{60}{\makebox[0pt][l]{\textbf{HuatuoGPT-o1-8B}}} & \rotatebox{60}{\makebox[0pt][l]{\textbf{FineMedLM-o1}}} & \rotatebox{60}{\makebox[0pt][l]{\textbf{DeepSeek-R1-Distill-Llama-8B}}} & \rotatebox{60}{\makebox[0pt][l]{\textbf{DeepSeek-R1-Distill-Qwen2.5-7B}}} & \rotatebox{60}{\makebox[0pt][l]{\textbf{GPT-4o}}} & \rotatebox{60}{\makebox[0pt][l]{\textbf{DeepSeek-V3}}} & \rotatebox{60}{\makebox[0pt][l]{\textbf{Qwen2.5-72B-Instruct}}} & \rotatebox{60}{\makebox[0pt][l]{\textbf{Llama3.3-70B-Instruct}}} & \rotatebox{60}{\makebox[0pt][l]{\textbf{BioMistral-7B}}} & \rotatebox{60}{\makebox[0pt][l]{\textbf{Llama-3.1-8B-UltraMedical}}} & \rotatebox{60}{\makebox[0pt][l]{\textbf{BioMistral-7B-SFT}}} \\ \hline
Bacteriology  &  \textbf{98.82} & 95.29 & 85.88 & 60.0 & 51.18 & 27.65 & 21.76 & 87.06 & 75.88 & 86.47 & 51.76 & 5.29 & 45.88 & 41.18 \\
Cardiology  &  \textbf{80.74} & 73.46 & 65.35 & 39.91 & 23.23 & 21.29 & 11.2 & 69.95 & 51.71 & 59.63 & 47.93 & 9.91 & 22.07 & 27.28 \\
Pediatric Cardiology  &  \textbf{81.08} & 71.62 & 62.76 & 35.93 & 23.88 & 15.22 & 11.63 & 67.07 & 54.18 & 54.29 & 46.57 & 9.49 & 22.02 & 19.62 \\
Dermatology  &  \textbf{80.74} & 71.48 & 62.22 & 32.96 & 19.63 & 18.15 & 7.22 & 62.78 & 47.04 & 55.0 & 40.93 & 12.41 & 20.0 & 28.52 \\
Endocrinology and Metabolism  &  \textbf{66.25} & 60.42 & 45.0 & 28.75 & 22.5 & 10.42 & 10.0 & 56.25 & 44.58 & 50.0 & 33.75 & 7.08 & 13.75 & 15.83 \\
Gynecology and Obstetrics  &  \textbf{81.01} & 70.2 & 64.9 & 43.96 & 23.36 & 20.27 & 12.55 & 68.72 & 48.46 & 62.75 & 51.95 & 16.85 & 26.44 & 26.64 \\
Genetics  &  \textbf{85.54} & 80.92 & 74.15 & 44.77 & 30.46 & 19.54 & 20.15 & 74.92 & 56.62 & 66.15 & 50.92 & 5.38 & 26.92 & 19.38 \\
Hematology  &  \textbf{86.42} & 78.54 & 67.59 & 39.42 & 20.15 & 17.59 & 9.27 & 69.85 & 60.22 & 59.12 & 49.56 & 12.7 & 22.34 & 24.96 \\
Hepato-Gastroenterology  &  \textbf{74.71} & 66.32 & 52.94 & 33.82 & 25.15 & 17.21 & 7.21 & 55.88 & 42.06 & 50.15 & 41.03 & 10.0 & 20.29 & 24.85 \\
Immunology  &  76.88 & \textbf{78.12} & 75.0 & 56.88 & 50.62 & 26.25 & 14.38 & 70.0 & 59.38 & 64.38 & 54.38 & 22.5 & 38.75 & 37.5 \\
Infectious Diseases  &  \textbf{73.75} & 70.68 & 57.39 & 37.5 & 22.39 & 14.89 & 8.86 & 62.73 & 44.32 & 53.18 & 40.57 & 11.36 & 17.39 & 22.39 \\
Forensic Medicine and Toxicology  &  \textbf{77.79} & 66.63 & 59.53 & 27.44 & 21.28 & 17.56 & 9.53 & 59.19 & 46.28 & 49.77 & 43.37 & 13.49 & 17.21 & 18.02 \\
Occupational Medicine  &  \textbf{54.38} & 50.0 & 38.75 & 22.5 & 16.88 & 11.25 & 9.38 & 43.75 & 32.5 & 30.0 & 30.62 & 10.62 & 8.75 & 19.38 \\
Neurology  &  \textbf{79.2} & 73.13 & 64.2 & 36.53 & 20.47 & 18.33 & 9.0 & 63.73 & 49.0 & 56.87 & 45.2 & 13.47 & 20.67 & 25.13 \\
Nephro-Urology  &  \textbf{83.58} & 75.85 & 66.11 & 42.02 & 23.26 & 18.13 & 10.62 & 70.1 & 54.09 & 59.84 & 48.39 & 11.14 & 24.77 & 26.84 \\
Ophthalmology  &  \textbf{77.1} & 69.03 & 56.13 & 41.94 & 23.23 & 19.68 & 6.45 & 60.32 & 46.13 & 62.26 & 40.0 & 11.29 & 21.94 & 25.81 \\
Orthopedics  &  \textbf{77.31} & 68.28 & 52.8 & 33.23 & 21.83 & 17.85 & 10.22 & 61.18 & 45.7 & 51.08 & 42.47 & 11.51 & 23.66 & 23.12 \\
Otorhinolaryngology (ENT)  &  \textbf{80.95} & 73.73 & 67.54 & 44.21 & 26.83 & 18.33 & 9.52 & 66.59 & 54.05 & 61.03 & 48.89 & 18.41 & 22.78 & 31.83 \\
Parasitology  &  \textbf{94.35} & 93.04 & 86.09 & 54.78 & 42.17 & 29.57 & 17.39 & 86.09 & 73.48 & 84.78 & 53.04 & 7.39 & 38.7 & 33.91 \\
Pharmacy  &  \textbf{86.51} & 79.29 & 71.96 & 43.67 & 30.98 & 19.36 & 9.07 & 75.04 & 54.4 & 65.79 & 49.68 & 17.03 & 26.56 & 25.51 \\
Pharmacology  &  68.89 & \textbf{75.56} & 62.22 & 35.56 & 18.89 & 32.22 & 8.89 & 62.22 & 44.44 & 48.89 & 26.67 & 10.0 & 11.11 & 11.11 \\
Pulmonology  &  \textbf{80.6} & 72.77 & 61.2 & 43.59 & 25.43 & 18.26 & 9.08 & 65.0 & 48.53 & 59.84 & 47.61 & 17.93 & 19.67 & 30.0 \\
Psychiatry  &  \textbf{78.5} & 69.12 & 62.38 & 36.88 & 21.25 & 22.0 & 9.12 & 63.62 & 46.0 & 60.0 & 47.12 & 14.62 & 20.25 & 30.12 \\
Pediatrics  &  \textbf{85.11} & 74.47 & 53.62 & 34.04 & 23.4 & 18.3 & 9.15 & 65.74 & 43.83 & 52.55 & 45.74 & 9.57 & 18.94 & 23.4 \\
Rheumatology  &  \textbf{75.14} & 65.14 & 58.38 & 35.41 & 25.41 & 16.22 & 8.11 & 58.92 & 46.22 & 48.92 & 37.03 & 9.73 & 17.3 & 24.05 \\
Intensive Care  &  \textbf{76.95} & 72.79 & 66.1 & 35.0 & 23.31 & 15.91 & 7.99 & 64.87 & 45.78 & 58.31 & 44.48 & 15.45 & 21.75 & 24.87 \\
Rehabilitation  &  \textbf{75.0} & 61.67 & 53.33 & 45.0 & 35.0 & 28.33 & 10.0 & 50.0 & 33.33 & 66.67 & 50.0 & 10.0 & 23.33 & 18.33 \\
Public Health  &  \textbf{71.05} & 64.74 & 43.68 & 36.32 & 32.63 & 27.37 & 18.42 & 50.53 & 39.47 & 52.11 & 40.0 & 18.42 & 22.63 & 21.58 \\
Semiology  &  \textbf{95.0} & \textbf{95.0} & 85.0 & 60.0 & 15.0 & 5.0 & 10.0 & 85.0 & 65.0 & 65.0 & 50.0 & 0.0 & 60.0 & 10.0 \\
Emergency Medicine  &  \textbf{74.13} & 65.05 & 59.82 & 33.49 & 22.94 & 16.88 & 8.35 & 58.53 & 42.48 & 54.04 & 43.49 & 11.83 & 20.92 & 21.93 \\ \hline
\end{tabular}
\caption{Performance by medical subject on the MediQAl-OEQ subset. The scores, obtained in zero-shot, are measured with LLM-as-judge.}
\label{tab:OEQ-subjects}
\end{table}

\end{document}